\pdfoutput=1

\documentclass[11pt]{article}

\usepackage[]{acl}
\usepackage{times}
\usepackage{latexsym}
\usepackage{enumitem} 
\usepackage[T1]{fontenc}

\usepackage[utf8]{inputenc}

\usepackage{microtype}

\usepackage{graphicx}
\usepackage{inconsolata}
\usepackage{xcolor}
\usepackage{calc}
\usepackage{float}
\usepackage{pifont}
\usepackage{tabularx}
\usepackage{booktabs}
\usepackage{amsmath}
\usepackage{amssymb}
\usepackage{tcolorbox}
\usepackage{url}
\usepackage{subfigure}
\usepackage{CJKutf8}
\newcommand{\chinese}[1]{\begin{CJK}{UTF8}{gbsn}#1\end{CJK}}
\usepackage{hyperref}
\usepackage{xurl}
\usepackage[frozencache,cachedir=minted-cache]{minted}
\newcommand{\OURS}{\textsc{$\infty$Bench}}%
\newcommand{\OURSSPACE}{\textsc{$\infty$Bench }}%

\definecolor{darkgreen}{rgb}{0.0, 0.5, 0.0} 
\definecolor{darkred}{rgb}{0.5, 0.0, 0.0}   

\newcommand{\cmark}{\textcolor{darkgreen}{\ding{51}}}%
\newcommand{\xmark}{\textcolor{red}{\ding{55}}}%

\title{\OURS: Extending Long Context Evaluation \\ Beyond 100K Tokens}

\author{Xinrong Zhang, Yingfa Chen, Shengding Hu, Zihang Xu, Junhao Chen\\
\textbf{Moo Khai Hao, Xu Han, Zhen Leng Thai, Shuo Wang, Zhiyuan Liu, Maosong Sun}\\
{Department of Computer Science and Technology, Tsinghua University, Beijing, China}\\
\texttt{zxr19@mails.tsinghua.edu.cn}}

\begin{document}
\maketitle
\begin{abstract}

Processing and reasoning over long contexts is crucial for many practical applications of Large Language Models (LLMs), such as document comprehension and agent construction. Despite recent strides in making LLMs process contexts with more than 100K tokens, there is currently a lack of a standardized benchmark to evaluate this long-context capability. Existing public benchmarks typically focus on contexts around 10K tokens, limiting the assessment and comparison of LLMs in processing longer contexts. In this paper, we propose \OURS, the first LLM benchmark featuring an average data length surpassing 100K tokens. \OURSSPACE comprises synthetic and realistic tasks spanning diverse domains, presented in both English and Chinese. The tasks in \OURSSPACE are designed to require well understanding of long dependencies in contexts, and make simply retrieving a limited number of passages from contexts not sufficient for these tasks. In our experiments, based on \OURS, we evaluate the state-of-the-art proprietary and open-source LLMs tailored for processing long contexts. The results indicate that existing long context LLMs still require significant advancements to effectively process 100K+ context. We further present three intriguing analyses regarding the behavior of LLMs processing long context. Our code and data is released\footnote{\url{https://github.com/OpenBMB/InfiniteBench}}\footnote{\url{https://huggingface.co/datasets/xinrongzhang2022/InfiniteBench}}.


\end{abstract}

\section{Introduction}
\begin{figure}
    \centering
    \includegraphics[width=\linewidth]{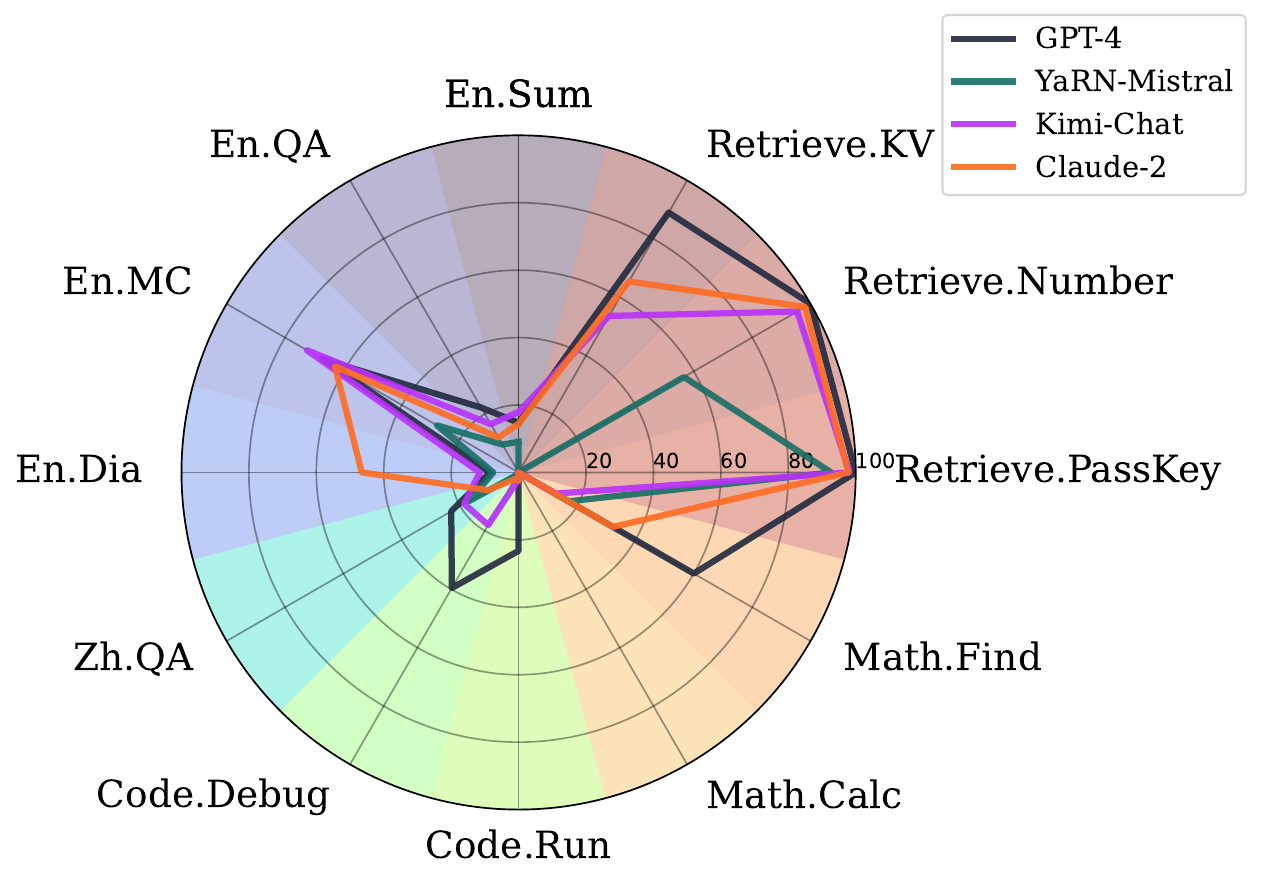}
    \caption{The performance of GPT-4, Kimi-Chat, YaRN-Mistral, and Claude 2 on \OURS. A higher value represents better performance.}
    \vspace{-0.3cm}
    \label{fig:benchmark-result}
\end{figure}
\begin{table*}[t]
    \footnotesize
    \centering
    \begin{tabular}{l|cccccccc}
        \toprule
        Benchmark  
            & Avg Len & En & Zh & Code & Math & Novel & Dialogue & Synthetic \\
        \midrule
        LRA~\citep{tay2020long}      
            & $\sim$10K
            & \cmark & \xmark & \xmark & \cmark & \xmark & \xmark & \cmark\\
        LongBench~\citep{bai2023longbench}       
            & $\sim$10K
            & \cmark & \cmark & \cmark & \xmark & \cmark & \cmark & \cmark\\
        L-Eval~\citep{An2023LEvalIS}     
            &  4K - 60K
            & \cmark & \xmark & \cmark & \cmark & \xmark & \xmark & \cmark\\
        LooGLE~\citep{Li2023LooGLECL}
            & $\sim$20K 
            & \cmark & \xmark & \xmark & \xmark & \xmark & \cmark & \xmark\\
        \OURSSPACE (ours) 
            & $\sim$200K
            & \cmark & \cmark & \cmark & \cmark & \cmark & \cmark & \cmark \\
        \bottomrule
    \end{tabular}
    \caption{Comparison to existing long-context benchmarks and \OURS. ``En'' and ``Zh'' refer to English and Chinese tasks. ``Code'', ``Math'', ``Novel'', ``Dialogue'' indicate whether the domain includes tasks from those domains, and ``Synthetic'' indicates whether there are auto-generated tasks.}
    \label{tab:table-benchmark-comparison}
\end{table*}

In recent years, large language models (LLMs) \cite{gpt3, gpt4, touvron2023llama} have exhibited exceptional performance across a range of natural language processing (NLP) tasks~\cite{qiu2020pre, han2021pre}. LLMs are showing a promising direction toward generalist task assistance, being capable of aiding users in practical tasks through conversational interactions. These tasks include web navigation~\cite{nakano2021webgpt}, analysis of code repositories~\cite{chen2021evaluating}, and extraction of useful information from documents~\cite{kovcisky2018narrativeqa}, indicating a step towards artificial general intelligence. For these LLM-based scenarios, the ability to process long contexts is increasingly critical, in addition to understanding fine-grained semantics and possessing extensive knowledge~\cite{dong2023survey, huang2023advancing}. Textual documents, historical dialogues, complex instructions, and cumbersome workflows, which constitute the data most directly processed in daily tasks, must be input to LLMs as long contexts for effective processing.

Despite this growing importance, LLMs consistently face challenges in processing long contexts, primarily due to the substantial computational resources required for long sequence training~\cite{dao2022flashattention, dao2023flashattention2} as well as the apparent inability to generalize to sequences longer than those encountered during training~\cite{chen2023extending,yarn}. 
LLMs are typically trained on sequences containing no more than 8K tokens~\cite{touvron2023llama, penedo2023refinedweb, biderman2023pythia}, and thus cannot well handle contexts exceeding 8K tokens.
These limitations have largely restricted most LLMs from being applied to more complex tasks.

Recent advancements in training infrastructure~\cite{Shoeybi2019MegatronLMTM,narayanan2021efficient, dao2022flashattention, dao2023flashattention2}, and efforts to improve length generalization~\cite{anil2022exploring,Chen2023ExtendingCW,yarn}\footnote{\url{https://www.reddit.com/r/LocalLLaMA/comments/14lz7j5/ntkaware_scaled_rope_allows_llama_models_to_have/}} have led to rapid developments in long-context LLMs. 
Based on these improved training infrastructures and length generalization methods, several LLMs have purportedly managed to process data exceeding 100K tokens~\citep{yarn,gpt4turbo,Yi-6B-200K,Yi-34B-200K}, with Claude 2~\cite{claude2} and Kimi-Chat~\cite{kimi} even claiming to be able to process up to 200K tokens.
However, the rapid emergence of long-context LLMs has outpaced the development of adequate evaluation benchmarks. Present long-context benchmarks predominantly feature contexts averaging around 10K tokens~\citep{bai2023longbench, tay2020long}, invariably falling below 100K tokens. \textbf{This lag in the advancement of long-context evaluation methodologies impedes both the comparative analysis of diverse long-context LLMs and the pinpointing of potential enhancements in long-context processing}.



In this work, we present \OURSSPACE, the first comprehensive benchmark featuring an average data length surpassing 100K tokens. \OURSSPACE includes tasks in different domains (novels, code, math, etc.) and languages (English and Chinese). To fully evaluate the performance of long-context LLMs, \OURSSPACE integrates synthetic tasks that can be auto-generated for even longer contexts (e.g., finding the top-$k$ number in an array)  in addition to a set of realistic tasks.

To construct tasks annotated by humans, we develop 5 annotation pipelines for detailed example annotation. These pipelines undergo iterative refinement until the examples meet quality standards. Auto-generated tasks, conversely, can be easily scaled to various lengths. Upon completing \OURS, we assess the performance of several state-of-the-art (SOTA) long-context LLMs on this benchmark to gauge its difficulty and evaluate the effectiveness of these models. The results show that current SOTA LLMs are not fully equipped to handle all tasks within \OURS, highlighting the ongoing challenge of enabling LLMs to process long contexts effectively. We also conduct intriguing analyses on the behavior of LLMs on such long contexts, including the task length ablation, the absent of ``lost in the middle phenomenon~\cite{lost-in-the-middle}'', and the context recalling prompting techniques.

Our contributions can be summarized as follows:
\begin{itemize}
    \item We construct and release \OURS, the first multi-domain bilingual benchmark for evaluating the ability to understand and reason over contexts surpassing 100K tokens. 
    \item We evaluate SOTA long-context LLMs on \OURS, which reveals severe performance degradation of these LLMs when scaling context lengths. These experimental results and analysis also indicate promising directions to improve long-context LLMs.
\end{itemize}

\section{Related Work}



\paragraph{Extending Context Length}

Transformers, typically trained on text sequences under 8K tokens due to self-attention's quadratic complexity, face challenges in longer downstream tasks. To address this, two main strategies have emerged: firstly, the development of positional encodings capable of handling longer text sequences~\cite{sun2022length, press2021train}, and secondly, the refinement of inference stage techniques to extend current LLMs post-training. The primary approach involves modifying rotary positional encoding~\cite{su2023roformer} and implementing post-training adjustments to better manage the increased relative positional distances in longer sequences~\cite{zhu2023pose, yarn, chen2023extending}.

\paragraph{100K+ LLMs}

Recently, many LLMs have shown the ability to handle over 100K tokens. Some popular proprietary 100K+ LLMs include GPT-4, Claude 2~\citep{claude2}, and Kimi-Chat~\citep{{kimi}}. 
On the other hand, there are much fewer open-source 100K+ models. Some notable models include YaRN~\citep{yarn} and Yi-200K~\citep{Yi-34B-200K, Yi-6B-200K}.
In this paper, we benchmark GPT-4, Claude 2, Kimi-Chat, and YaRN-Mistral-7B-128K\footnote{\url{https://huggingface.co/NousResearch/Yarn-Mistral-7b-128k}, we denote this model by YaRN-Mistral.} on \OURS, which are some of the latest and strongest LLMs that claim to be able to handle over 100K tokens. 

\paragraph{Inference Infrastructure}
Numerous studies aim to accelerate self-attention computation. Research primarily concentrates on refining attention mechanisms through improved IO management~\citep{dao2022flashattention,dao2023flashattention2}, memory optimization~\citep{kwon2023efficient, Shazeer2019FastTD, Ainslie2023GQATG}, and enhanced parallelization in decoding~\citep{flash-decoding,flash-decoding-plusplus}. Approaches like Sliding Window Attention~\citep{beltagy2020longformer}, LM-Infinite~\citep{Han2023LMInfiniteSO}, and StreamingLLM~\citep{Xiao2023EfficientSL} introduce attention variants for handling infinitely long sequences without overwhelming computation or memory overhead. However, these techniques often face challenges in maintaining historical information.

\paragraph{Long Context Benchmarks}

Several benchmarks exist for evaluating long-context AI models, notably featuring context lengths of around 10K tokens. L-Eval~\citep{An2023LEvalIS} and LongBench~\citep{bai2023longbench} are prominent examples, aggregating pre-existing tasks\cite{Kocisk2017TheNR,Dasigi2021ADO,Yang2018HotpotQAAD,Huang2021EfficientAF,Joshi2017TriviaQAAL} into comprehensive benchmarks. LongBench encompasses four categories—QA, summarization, synthetic retrieval, and code—spanning 21 tasks, with four being novel. Conversely, L-Eval incorporates 18 tasks across QA, summarization, math, retrieval, and multiple-choice (MC) domains, introducing three new tasks. Another notable benchmark, LooGLE~\citep{Li2023LooGLECL}, differentiates between short and long dependency examples, focusing on summary and QA tasks; its summary corpus contrasts with ours, utilizing academic papers over novels. The Long-Range Arena (LRA) \citep{tay2020long} further diversifies with six tasks in text, image, and math, designed for scalability. In comparison, \OURSSPACE stands out for its substantially longer contexts and a broader range of task domains. Table~\ref{tab:table-benchmark-comparison} offers a detailed comparison of these long-context benchmarks.

\begin{figure}[!t]
    \centering
    \includegraphics[width=\linewidth]{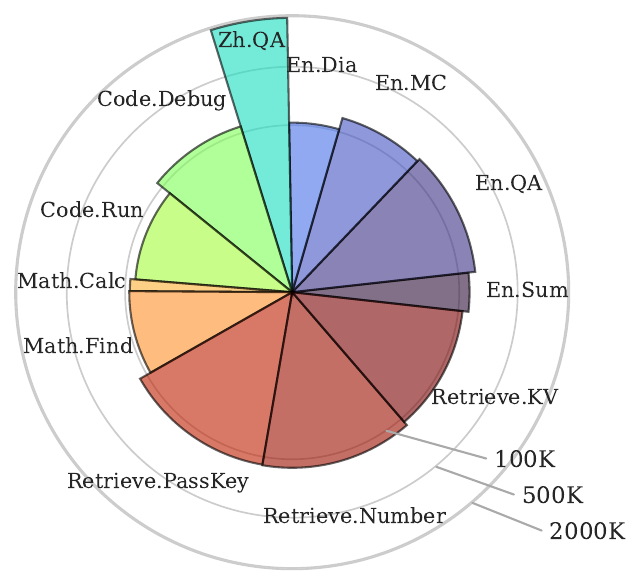}
    \caption{The statistics of the data in \OURS. The radius of each segment indicates the length of input plus output on the logarithmic scale, and the width (or angle) indicates the number of examples (proportionally to the total number of examples).}
    \label{fig:data-distribution}
\end{figure}

\section{\OURS} 

\begin{table}[!t]
    \footnotesize
    \centering
    \begin{tabular}{l|cccc}
        \toprule
        \textbf{Task}  & \textbf{Annotation}  & \textbf{\# Ex.} & \textbf{Avg Len} \\
        \midrule
        Ret.PassKey    & Auto & 590 & 122.4K/2\\
        Ret.Number     & Auto & 590 & 122.4K/4\\
        Ret.KV         & Auto & 500 & 121.1K/22.7\\
        \midrule
        En.Sum      & Human & 103 & 103.5K/1.1K\\
        En.QA       & Human & 351 & 192.6k/4.8\\
        En.MC       & Human & 229 & 184.4K/5.3\\
        Zh.QA       & Human & 189 & 2068.6K/6.3 \\
        En.Dia      & Auto & 200 & 103.6K/3.4 \\
        \midrule
        Code.Debug  & Human & 394 & 114.7K/4.8\\
        Code.Run    & Auto & 400 & 75.2K/1.3 \\
        \midrule
        Math.Calc   & Auto & 50 & 43.9K/43.9K \\
        Math.Find   & Auto & 350 & 87.9K/1.3 \\
        \bottomrule
    \end{tabular}
    \caption{Data statistics. The columns indicate whether the annotation was auto-generated or done by humans, the number of examples, and the average length (input/output) in tokens.}
    \label{tab:data-statistics}
\end{table}
\begin{table*}[!ht]
    \centering
    \begin{tabular}{l|cccc}
        \toprule
        \textbf{Task}    
            &  \textbf{GPT-4}    
            &  \textbf{YaRN-Mistral}   
            &  \textbf{Kimi-Chat}    
            &  \textbf{Claude 2}   \\
        \midrule
        Retrieve.PassKey     
            & \textbf{100.00}            
            & 92.71              
            & 98.14        
            & 97.80          \\
        Retrieve.Number      
            & \textbf{100.00}            
            & 56.61              
            & 95.42        
            & 98.14          \\
        Retrieve.KV          
            & \textbf{89.00}          
            & 0.00   
            & 53.60        
            & 65.40       \\  
        En.Sum               
            & 14.73           
            & 9.09   
            & \textbf{17.93}         
            & 14.45          \\
        En.QA                
            & \textbf{22.22}          
            & 9.55              
            & 16.52        
            & 11.97          \\
        En.MC                
            & 67.25         
            & 27.95              
            & \textbf{72.49}        
            & 62.88         \\
        En.Dia               
            & 8.50           
            & 7.50               
            & 11.50        
            & \textbf{46.50}  \\
         Zh.QA                
            & \textbf{23.06}          
            & 16.98              
            & 18.62        
            & 10.53    \\
         Code.Debug           
            & \textbf{39.59}          
            & 0.76   
            & 18.02       
            & 2.28   \\
         Code.Run             
            & \textbf{23.25}          
            &  1.25   
            &  2.00   
            &  2.50   \\
         Math.Calc            
            & \textbf{0.01} 
            & 0.00  
            & 0.00  
            & 0.00   \\
         Math.Find            
            & \textbf{60.00}          
            & 17.14              
            & 12.57        
            & 32.29    \\
        \midrule
        Average
        & \textbf{45.63}
        & 19.96
        & 34.73
        & 37.06\\
        \bottomrule
    \end{tabular}
    \caption{Main results. The performance of the baselines in \OURS. For multiple-choice questions, if the model does not output one of the options, we regard it as an empty prediction, and thus give it a score of 0.}
    \label{tab:main-results}
\end{table*}

\OURSSPACE encompasses 12 tasks spanning 5 domains: retrieval, code, math, novels, and dialogue. Two of these tasks are derived from existing literature\cite{Mohtashami2023LandmarkAR,lost-in-the-middle}. Among the newly introduced tasks, half are generated automatically, while the remainder are annotated by humans. 

In total, \OURSSPACE includes 3946 examples, featuring a length beyond 100K tokens (average approximately 200K). Figure~\ref{fig:data-distribution} illustrates the distribution of these tasks. Table~\ref{tab:data-statistics} details their respective input and output lengths as well as the number of examples per task. 

Next, we illustrate each task in detail. The tasks can be grouped into two broad categories. The first involves realistic context collected from real-world scenarios which has potential practical usage of long context LLMs. The second depends on synthetic contexts which are created or collected for testing certain capabilities of long-context LLMs. 

\subsection{Realistic Context}

\subsubsection{Novel}

\begin{figure}[t]
    \centering
    \includegraphics[width=\linewidth]{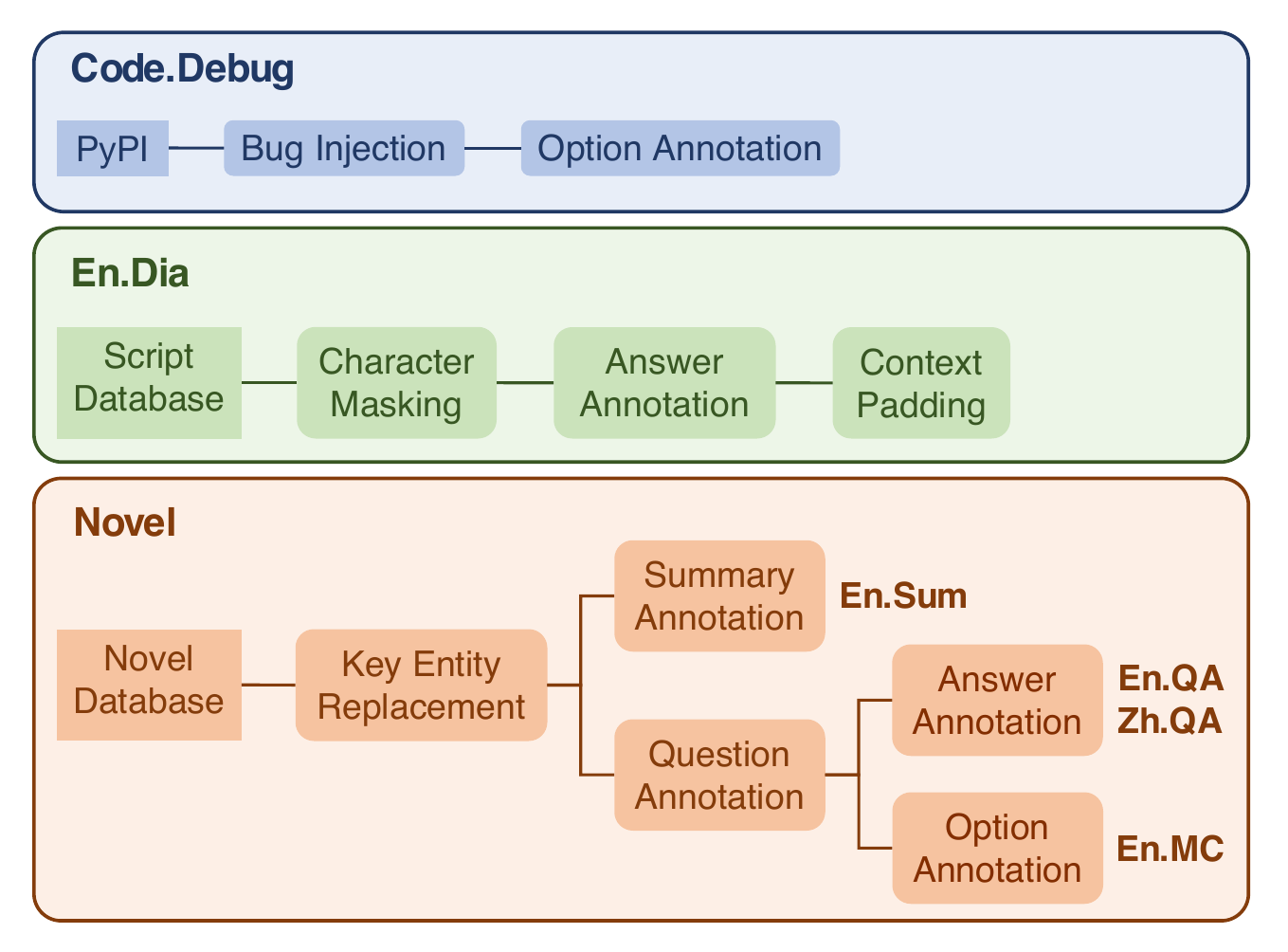}
    \caption{The annotation pipelines for the human-annotated tasks in \OURS.}
    \label{fig:annotation-pipeline}
\end{figure}

We develop novel-based tasks as outlined in Figure~\ref{fig:annotation-pipeline}, utilizing novels sourced from websites\footnote{\label{SparkNotes}\url{https://www.sparknotes.com/}}\footnote{\label{CliffsNotes}\url{https://www.cliffsnotes.com/}} and are manually filtered. More annotation information in Appendix.~\ref{sec:annotation-process}.

In these tasks, models are tasked with reasoning over entire novels presented during inference. Recognizing that many novels, along with their movie adaptations and related discussions, are accessible online and may have been encountered by LLMs during training, we adopt \textit{key entity replacement} as a countermeasure. This involves substituting prominent entities determined by annotators, such as main character names, with unrelated ones, creating ``fake novels''.

Using these altered novels, we design tasks in three formats: summarization, open-form question answering (QA), and multiple-choice (MC) questions, applying key entity replacement to the annotations as well. All English tasks share the same set of modified novels.




\paragraph{En.Sum} 

The En.Sum task requires models to generate a concise summary of the novel. Gold standard labels are sourced from the web and undergo manual filtering to remove non-summarization content, like comments. Model performance is evaluated using the ROUGE-L-Sum metric~\citep{lin-2004-rouge}.


\paragraph{En.QA \& Zh.QA} 

We employ the same annotation pipeline for both En.QA and Zh.QA tasks, ensuring that the questions necessitate long-range dependency and reasoning, beyond simple short passage retrieval. The tasks are primarily categorized into two types of reasoning:
\begin{itemize}
    \item Aggregation: This involves compiling various pieces of information scattered throughout the novel. An example question in \OURSSPACE is ``How much money in total did A spend on lunch?''
    \item Filtering: This requires identifying specific information from a larger set. An example question in \OURSSPACE is ``What color dress did A wear when A met B for the second time?''
\end{itemize}

These tasks test LLMs to locate and process information within the novel, performing reasoning through aggregation or filtering to derive answers.

\paragraph{En.MC} 

The En.MC task is annotated similarly to En.QA, but differs in that the model is presented with four answer choices. Annotators are instructed to craft these options to be challenging.


\subsubsection{Dialogue}

\paragraph{En.Dia}

The construction process for the En.Dia task is depicted in Figure~\ref{fig:annotation-pipeline}. We gather movie and drama scripts from a designated online database\footnote{\url{https://imsdb.com/}}, focusing on a corpus of long, multi-role dialogues. Only the English scripts are retained and necessary cleaning is applied.

In the En.Dia task, random instances of character names within a script are replaced with \textcolor{blue}{\texttt{\$\$MASK\$\$}}. The objective is to correctly identify these masked names. For scripts falling short of 100K tokens, we augment them by padding with additional scripts.




\subsubsection{Code}

\paragraph{Code.Debug} 

We develop the task as per the process illustrated in Figure~\ref{fig:annotation-pipeline}. Code repositories, sourced from PyPI\footnote{\label{PyPI}\url{https://pypi.org/}}, undergo a filtering process, and those outside the 64K to 256K token range are excluded (tokenization via the tiktoken tokenizer\cite{Tiktoken}). Each repository is transformed into a single file, aggregating the content from all files within, each prefaced by its relative path to the root directory. Three of the authors then insert a deliberate and obvious error into one function per repository. The options are presented in the \texttt{Class.Function} format. Six methods are employed for bug insertion: (1) deleting a necessary variable declaration; (2) using an incorrect number of arguments in function calls; (3) creating infinite loops; (4) causing indentation errors; (5) substituting references with undefined variable/function names; (6) introducing blatant syntax errors (e.g., non-closed brackets).

Initial results indicate that this task is too challenging for current LLMs (None of the baseline models can identify the most obvious error such as non-closed brackets). To mitigate this, we offer four answer choices, one containing the injected bug and the others are bug-free. Note that this makes many examples easily solved by external retrieval preprocess. However, we encourage the users not to use external retrieval preprocess to keep the evaluation fair. And we are looking forward to the stage where LLMs can directly solve the problem without selection options.

\subsection{Synthetic Context}

The second category of tasks is characterized by a synthetic context. These tasks, devoid of direct real-world application or use case, are engineered to evaluate the capability for processing lengthy contexts. We delineate four essential ability for effective long-context processing:

\begin{enumerate}[itemsep=0pt, parsep=0pt]
    \item Location and retrieval. This encompasses all retrieval tasks.
    \item Elevated information resolution. This involves the Retrieve.Number task.
    \item State preservation. This incorporates the Code.Run and Math.Find functions.
    \item Sequential processing. This utilizes the Math.Calc function.
\end{enumerate}

\subsubsection{Retrieve}

In retrieval tasks, models retrieve specific character sequences from lengthy contexts with predominantly irrelevant content. Such tests, adaptable for any context length, can assess the impact of information placement on model performance, like the \emph{lost-in-the-middle} phenomenon~\citep{lost-in-the-middle}. The three retrieval tasks in \OURSSPACE vary in complexity.


\paragraph{Retrieve.PassKey} 

This task is first proposed by \citet{Mohtashami2023LandmarkAR}. Models are prompted to find a specific \textcolor{blue}{\texttt{<key>}} called pass key, which is a random 5-digit sequence. The pass key is inserted into a lengthy and noisy context, as shown below. In \OURS, we generate examples with 59 different pass key locations that are evenly distributed in the context. At each location, we construct 10 examples with different pass keys. This results in 590 examples.
\begin{tcolorbox}
\small
There is an important pass key hidden in a lot of irrelevant text. Find it.\\
\textcolor{orange}{<very long noisy context>}\\
The pass key is \textcolor{blue}{<key>}. Remember it. The pass key is \textcolor{blue}{<key>}\\
\textcolor{orange}{<very long noisy context>}\\
What is the pass key?
\end{tcolorbox}

\paragraph{Retrieve.Number} 

To examine the local attention of LLMs, we have enhanced the complexity of Retrieve.PassKey by increasing the answer length to 10 digits and \textit{incorporating successive repetitive digits}. For example, a \textcolor{blue}{\texttt{<key>}} in Retrieve.PassKey valued \texttt{98762}, while in Retrieve.Number is \texttt{9998877762}. This modification aims to assess the local resolution capabilities of long context models, as our preliminary experiments indicate that LLMs struggle with discerning repeated numbers.

\paragraph{Retrieve.KV} 

\citet{lost-in-the-middle} introduce a key-value retrieval task within a large JSON object containing many key-value pairs (e.g., \texttt{30eea139-b6dd-43fc-bc5d-0d3d17980229} $\rightarrow$ \texttt{bfd36c2b-c57e-41ef-9cc1-b21b4e60e664}). This task demands the model to accurately identify and retrieve the value corresponding to a specified key. The complexity of this task is heightened due to the indistinguishable format of relevant and irrelevant information.

\subsubsection{Code}
\paragraph{Code.Run} 

In this task, we evaluate the ability of LLMs to simulate multi-step function executions that involve basic arithmetic operations. While this task is readily solvable using a Python interpreter, the focus here is on the long-term state tracking required in such tasks. The capability of state tracking has been demonstrated in GPT-4~\citep{bubeck2023sparks}. Specifically, the task involves creating Python code consisting of multiple simple functions, incorporating operations such as addition, subtraction, and nested function calls. The structural design of these tasks is as follows:

\begin{minted}{python}
def func_0(x):
    return func_1(x) + 4

def func_1(x):
    return x - 1
\end{minted}


Some functions' return values are dependent on other functions (e.g., \textcolor{blue}{\texttt{func\_0}} invokes \textcolor{blue}{\texttt{func\_1}}). We define \emph{depth} as the number of cascading function calls initiated by a single call. Thus, the depth for \textcolor{blue}{\texttt{func\_1}}'s call within \textcolor{blue}{\texttt{func\_0}} is 1. In Code.Run, we employ depths ranging from 2 to 10, ensuring each function calls at most one other function. To keep the simplicity of each single step of computation, these functions are restricted to performing only addition and subtraction.


\subsubsection{Math}


\paragraph{Math.Find} 

Math.Find assesses the model's capability to identify specific elements within a large array, requiring comprehensive observation for accuracy. This task also tests the ability to preserve states while encoding the context. Concretely, the model receives an extensive list of numbers and is tasked with locating one of seven key numbers: the three largest (1st, 2nd, and 3rd), the three smallest (1st, 2nd, and 3rd), and the median.

\paragraph{Math.Calc} 

To assess sequential processing skills, Math.Calc prompts the model to compute the result of a lengthy arithmetic expression featuring addition and subtraction. Initial experiments indicate that current LLMs struggle to directly produce the final answer. Hence, we instead query the LLMs to provide the intermediate result following each operator. Model performance is evaluated based on the number of correct values preceding the first error.

\section{Experiments} 

We conduct a thorough set of experiments on \OURS. We will introduce the baselines, experimental setup, and main results in this section.
\subsection{Baselines}
\OURSSPACE generally requires the ability to handle input contexts longer than 100k. There is a handful of LLMs that claim to be capable of handling contexts over 100k. We include four baselines. The first three are proprietary LLMs as we do not have access to the model, while the last baseline is open-sourced. Details on evaluation are in Appendix.~\ref{sec:evaluation-process}.

\paragraph{GPT-4} 
GPT by OpenAI is one of the most widely used and capable LLMs in the market, and a recent version of GPT-4 ~\citep{gpt4turbo} can support 128K contexts. 

\paragraph{Claude 2}
Claude 2~\citep{claude2} is a proprietary chat-based LLM released by Anthropic AI and has shown impressive capabilities. The second version of the Claude series supports 200K contexts. We manually enter each example through the webpage because we have no access to their API.

\paragraph{Kimi-Chat}
Kimi-Chat, a proprietary chat-oriented LLM developed by Moonshot AI \cite{kimi}, is designed to process contexts up to 200K. Due to the lack of API access, we manually input the test data using their web interface.


\paragraph{YaRN-Mistral}
YaRN-Mistral is a derivative of Mistral-7B~\citep{mistral} introduced by \citet{yarn}. The original Mistral-7B was trained on input lengths up to 8K and shows a reduced performance in longer contexts. \citet{yarn} adapted it to 128K contexts by modifying the position encoding and continued training.


\subsection{Experimental Setup}
\paragraph{Prompt Templates}

For each model-task combination, we craft prompts to optimize model performance on short dummy examples. Detailed prompt templates for each model and task can be found in Appendix~\ref{sec:prompt-templates}.

\paragraph{Input Truncation}
All API-based baselines are subject to a maximum input length limit and will reject inputs exceeding this threshold. While YaRN-Mistral is theoretically capable of handling longer contexts, the authors only claim abilities up to 128K. Therefore, inputs are truncated by removing the center and joining both ends. This approach is predicated on the assumption that key information, such as instructions and book titles, is typically located at either the start or the end of a prompt.


\subsection{Main Result}
\begin{figure*}[!ht]
    \centering
    \includegraphics[width=\linewidth]{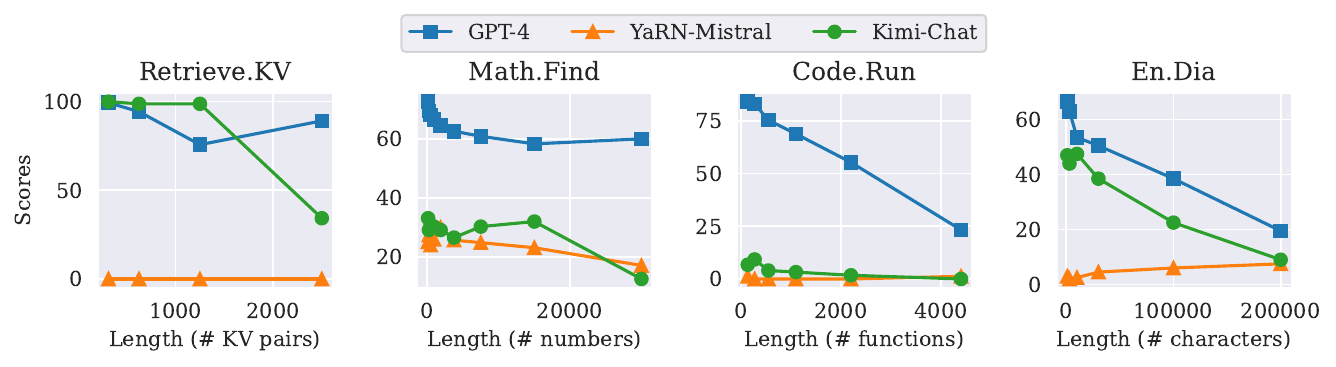}
    \caption{Baseline performance as a function of input length.}
    \label{fig:score-by-length}
\end{figure*}

Table~\ref{tab:main-results} and Figure~\ref{fig:benchmark-result} display the performances of various baselines on \OURS. Notably, GPT-4 outperforms other baselines in the retrieval, code, and math domains, with a considerably higher average score. However, in the novel-based tasks, no distinct winner emerges among the proprietary LLMs. On the other hand, the open-source YaRN-Mistral lags behind the proprietary models in most tasks, exhibiting almost random performance in multiple areas. This aligns with its relatively inferior performance in shorter contexts compared to these models. Additionally, it is observed that the baselines generally excel more in retrieval tasks than in other areas, echoing the relative simplicity of these tasks for human participants.

\section{Analysis}

We subsequently perform a detailed analysis of the results, identifying and emphasizing several notable and interesting phenomena.


\subsection{Length Ablation}

In line with our benchmark's goal to assess proficiency in managing lengthy contexts, we verify the baselines' capability with shortened context versions. A subset of the auto-generated tasks is modified accordingly, and the performance outcomes are illustrated in Figure \ref{fig:score-by-length}. It is observed that model performance generally declines with longer input lengths compared to shorter ones. This suggests that while these baselines are technically equipped to handle extended inputs, their effectiveness diminishes significantly under such conditions.

\subsection{Lost in the middle}
\begin{figure*}[!ht]
    \centering
    \includegraphics[width=0.9\linewidth]{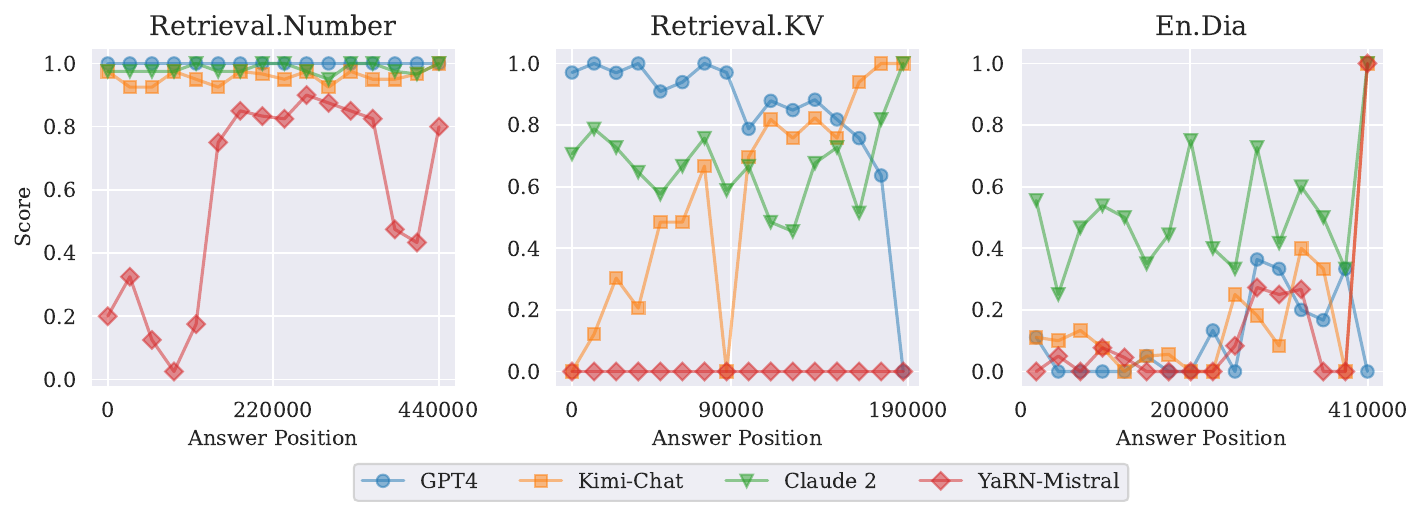}
    \caption{Performance as a function of the answer position (in the number of characters). The steep drop in performance for Kimi-Chat in the middle on Retrieval.KV is caused by the answer being removed by truncation.}
    \label{fig:performance-by-answer-pos}
\end{figure*}

\begin{figure}[!t]
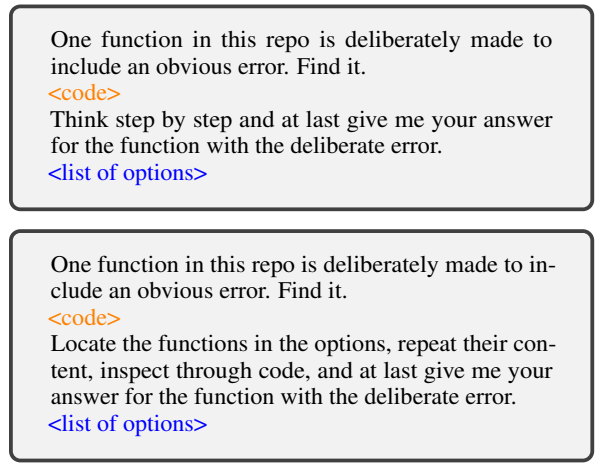

\begin{tcolorbox}
\small
One function in this repo is deliberately made to include an obvious error. Find it.\\
\textcolor{orange}{<code>}\\
Think step by step and at last give me your answer for the function with the deliberate error.\\
\textcolor{blue}{<list of options>}
\end{tcolorbox}

\begin{tcolorbox}
\small
One function in this repo is deliberately made to include an obvious error. Find it.\\
\textcolor{orange}{<code>}\\
Locate the functions in the options, repeat their content, inspect through code, and at last give me your answer for the function with the deliberate error.\\
\textcolor{blue}{<list of options>}
\end{tcolorbox}
\caption{Compared to the first prompt, the second prompt improves GPT-4's results on Code.Debug dramatically.}
\end{figure}
Prior research indicates a performance decline in some LLMs when answers are positioned around the center of the context~\citep{lost-in-the-middle}. However, our findings do not strongly corroborate this. As depicted in Figure~\ref{fig:performance-by-answer-pos}, we analyze model performance based on answer location in three location-dependent tasks. We observe \textit{no consistent trend between performance and answer position across different models}. For instance, GPT-4 shows a preference for early answers in Retrieval.KV but favors later ones in En.Dia. In contrast, Claude 2's performance remains relatively unaffected by answer position on all three tasks, whereas YaRN-Mistral and Kimi-Chat excel with end-positioned answers (except that YaRN-Mistral get zero performance on all positions on Retrieval.KV).



One plausible reason why we have different observations from \citet{lost-in-the-middle} is that they experiment with different models using at most 16K length contexts, which is about 8 times shorter than our setting. The models in their study are also different from ours.
Finally, the tasks are different: their experiments involve document question answering (and their result with Retrieval.KV arguably does not show a very pronounced performance drop as well). We hypothesize that the phenomenon of ``Lost in the middle'' is only exhibited on specific tasks and models. A more thorough investigation of these differences is beyond the scope of this paper.


\subsection{Context Recalling}
\label{sec:pipeline-instruction}

We identify an intriguing prompting technique for tasks involving extended context, termed \textit{context recalling}. This technique posits that, although the information is present in the context and accessible via direct attention, it may be more effective to first prompt the model to \textit{recall the relevant information in its generation before engaging in further reasoning}. In our experiments using Code.Debug, when we merely instructed GPT-4 to process information step-by-step, the accuracy was \textbf{15.74\%}. However, by explicitly directing GPT-4 to repeat the relevant code before analysis, its accuracy on Code.Debug markedly improved to \textbf{39.59\%}. This approach of context recalling warrants additional investigation.


\section{Conclusions}

We introduce \OURS, the first benchmark tailored for long contexts exceeding 100K in average length. Empirical evidence indicates that despite claims of proficiency with such extensive contexts, current LLMs demonstrate significant performance degradation when dealing with them. This finding highlights the need for advanced methodologies to improve LLMs' efficiency in processing long context. Additionally, our analysis offers insights into LLM behavior in long-context tasks, guiding future research.





\newpage

\section*{Limitations}

While our benchmark offers valuable insights into LLM performance, it may not be sufficiently diverse or extensive to provide a comprehensive assessment of model capabilities, a constraint common to most benchmarks. Additionally, the reliance on exact match for scoring, dependent on prompt templates and answer parsing methods, may necessitate tailored redesigns for new model evaluations.

Furthermore, supporting contexts up to 100K tokens may fall short for applications requiring analysis of extensive datasets, such as multiple books or entire databases. Exploring LLMs' capacity to handle up to a million tokens or more presents a promising research avenue. In practical applications, finetuning models to memorize context, rather than processing it during inference, could offer a more efficient alternative, albeit with significant computational demands.



\section*{Ethics Statement}
Our human annotators are directed to exclude data that may raise sensitive ethical issues, such as offensive language or social biases. Nonetheless, the potential for encountering sensitive content persists, particularly if the sourced books or code contain such material. This concern is somewhat mitigated since the benchmark's primary focus is on evaluating the long-context capabilities of LLMs, rather than influencing their social bias.

The goal of this research is to advance the development of LLMs' proficiency in handling extensive contexts. This could aid in implementing more effective ``guardrails'' against misuse by incorporating detailed specifications prior to user interactions. However, this approach also potentially increases the risk of novel prompt injection attacks.



\newpage

\bibliography{acl_latex}

\appendix

\section{RWKV}

RWKV~\citep{rwkv} is an architecture that combines the power of the transformer architecture~\citep{transformer} and recurrent neural network~\citep{lstm}. Its training process can be parallelized while the inference procedure is recurrent, enabling $O(1)$ complexity during inference. Hence, the memory usage does not scale with context length, allowing it to support arbitrary-length inputs. We use the RKWV-4-World-7B version of this model series. However, we should keep in mind that this model was not trained on inputs of this length.

\begin{table}[!ht]
    \centering
    \begin{tabular}{l|c}
        \toprule
        \textbf{Model} &
        \textbf{Retrieve.PassKey Acc.}  \\ 
        \midrule
        GPT-4 Turbo     & \textbf{100\%}    \\
        YaRN-Mistral    & 92.71\%  \\            
        Kimi-Chat       & 98.14\%   \\     
        Claude 2        & 97.80\%        \\  
        RWKV-4-World-7B & 0.00\%  \\
        \bottomrule
    \end{tabular}
    \caption{Results in Retrieve.PassKey with RWKV-4-World-7B. Since RWKV-4 was only trained on 4k sequences, it has zero performance on \OURS. It outputs only unintelligible content in this test.}
    \label{tab:rwkv-results}
\end{table}

Table~\ref{tab:rwkv-results} shows the performance of RWKV-4-World-7 in comparison to our baselines. We find that RWKV-4-World-7B outputs unintelligible texts on our benchmark, which causes it to achieve zero performance on Retrieve.PassKey, which is the easiest task for other baselines. This is likely because this model was not trained on inputs of this length and suffers from train-test domain shift.\footnote{We emphasize that this result is not evidence that the architecture of RWKV is incapable of handling lengthy inputs.} Therefore, we do not consider testing it on other tasks in our benchmark.

\section{Prompt Templates}

In the following templates, many tasks has an \textcolor{blue}{<input>} part that is provided in each example. Generally, they are a short question-like text that tells the model what it is supposed to do. One example is ``What is the pass key?''.

\label{sec:prompt-templates}

\subsection{Retrieve.PassKey}
The prompt below applies to GPT-4, Claude 2, and Kimi-Chat.
\begin{tcolorbox}
\small
There is an important info hidden inside a lot of irrelevant text. Find it and memorize them. I will quiz you about the important information there.\\
\\
\textcolor{blue}{<context>}\\
\\
\textcolor{blue}{<input>}
\end{tcolorbox}

The prompt below applies to YaRN-Mistral.
\begin{tcolorbox}
\small
There is an important info hidden inside a lot of irrelevant text. Find it and memorize them. I will quiz you about the important information there.\\
\\
\textcolor{blue}{<context>}\\
\\
\textcolor{blue}{<input>}\\
\\
The pass key is
\end{tcolorbox}

\subsection{Retrieve.Number}
The prompt below applies to GPT-4, Claude 2, and Kimi-Chat.
\begin{tcolorbox}
\small
There is an important info hidden inside a lot of irrelevant text. Find it and memorize them. I will quiz you about the important information there.\\
\\
\textcolor{blue}{<context>}\\
\\
\textcolor{blue}{<input>}
\end{tcolorbox}

The prompt below applies to YaRN-Mistral.
\begin{tcolorbox}
\small
There is an important info hidden inside a lot of irrelevant text. Find it and memorize them. I will quiz you about the important information there.\\
\\
\textcolor{blue}{<context>}\\
\\
\textcolor{blue}{<input>}\\
\\
The sequence of digits is
\end{tcolorbox}

\subsection{Retrieve.KV}

\begin{tcolorbox}
\small
Extract the value corresponding to the specified key in the JSON object below.\\
\\
\textcolor{blue}{<context>}\\
\\
\textcolor{blue}{<input>}
\end{tcolorbox}

\subsection{En.Sum}

The prompt below applies to GPT-4, Claude 2, and Kimi-Chat.
\begin{tcolorbox}
Summarize the book below.\\
\\
\textcolor{blue}{<context>}
\end{tcolorbox}
The prompt below applies to YaRN-Mistral.
\begin{tcolorbox}
Summarize the book below.\\
\\
\textcolor{blue}{<context>}\\
\\
Summary:
\end{tcolorbox}

\subsection{En.QA}
The prompt below applies to GPT-4, Claude 2, and Kimi-Chat.
\begin{tcolorbox}
\small
Read the book below and answer a question.\\
\\
\textcolor{blue}{<context>}\\
\\
Question: \textcolor{blue}{<question>}\\
\\
Be very concise.
\end{tcolorbox}

The prompt below applies to YaRN-Mistral.
\begin{tcolorbox}
\small
Read the book below and answer a question. Be very concise in your answer.\\
\\
\textcolor{blue}{<context>}\\
\\
Question: \textcolor{blue}{<question>}\\
\\
Answer:
\end{tcolorbox}

\subsection{En.MC}
The prompt below applies to GPT-4, Claude 2, and Kimi-Chat.
\begin{tcolorbox}
\small
Read the book and answer the question.\\
\\
\textcolor{blue}{<context>}\\
\\
Question: \textcolor{blue}{<question>}\\
\\
Only one of the following options is correct, tell me the answer using one single letter (A, B, C, or D). Don't say anything else.\\
\\
A. \textcolor{blue}{<OPTION\_A>}\\
B. \textcolor{blue}{<OPTION\_B>}\\
C. \textcolor{blue}{<OPTION\_C>}\\
D. \textcolor{blue}{<OPTION\_D>}
\end{tcolorbox}

The prompt below applies to YaRN-Mistral.
\begin{tcolorbox}
\small
Read the book and answer the question.\\
\\
\textcolor{blue}{<context>}\\
\\
Question: \textcolor{blue}{<question>}\\
\\
Only one of the following options is correct, tell me the answer using one single letter (A, B, C, or D). Don't say anything else.\\
\\
A. \textcolor{blue}{<OPTION\_A>}\\
B. \textcolor{blue}{<OPTION\_B>}\\
C. \textcolor{blue}{<OPTION\_C>}\\
D. \textcolor{blue}{<OPTION\_D>}\\
\\
The correct option is: 
\end{tcolorbox}

\subsection{En.Dia}
The prompt below applies to GPT-4, Claude 2, and Kimi-Chat.
\begin{tcolorbox}
\small
Below is a dialogue script where one random occurrence of a character name is replaced with \$\$MASK\$\$, and you should try to guess who that character is.\\
\\
The dialogue:\\
\\
---\\
\\
\textcolor{blue}{<context>}\\
\\
---\\
\\
End of dialogue.\\
\\
Which character is most likely \$\$MASK\$\$? Just say the name used by the scriptwriter (before the colon marks) of one single character and nothing else.
\end{tcolorbox}

The prompt below applies to YaRN-Mistral.
\begin{tcolorbox}
\small
Below is a dialogue script where one random occurrence of a character name is replaced with \$\$MASK\$\$, and you should try to guess who that character is.\\
\\
The dialogue:\\
\\
---\\
\\
\textcolor{blue}{<context>}\\
\\
---\\
\\
End of dialogue.\\
\\
The name that has been replaced with \$\$MASK\$\$ is likely:
\end{tcolorbox}

\subsection{Zh.QA}
The prompt below applies to GPT-4, Claude 2, and Kimi-Chat.
\begin{tcolorbox}
\small
\chinese{请根据以下书籍回答我的问题。}(Read the book and answer the question.)\\
\\
\textcolor{blue}{<context>}\\
\\
\chinese{问题：}(Question: )\textcolor{blue}{<question>}\\
\\
\chinese{请尽量简短地回答。}(Be very concise.)
\end{tcolorbox}

The prompt below applies to YaRN-Mistral.
\begin{tcolorbox}
\small
\chinese{请根据以下书籍回答我的问题。}(Read the book and answer the question.)\\
\\
\textcolor{blue}{<context>}\\
\\
\chinese{问题：}(Question: )\textcolor{blue}{<question>}\\
\chinese{答案：(Answer:)}
\end{tcolorbox}

\subsection{Code.Debug}
The prompt below applies to GPT-4, Claude 2, and Kimi-Chat.
\begin{tcolorbox}
\small
There is ONLY ONE function in the large project that is deliberately made to include an obvious error. Please find the function that contains the most obvious errors. I will give you four options to narrow your scope. You can inspect through the options and think. Eventually, tell me the answer using one single letter (A, B, C, or D).\\
\\
\textcolor{blue}{<context>}\\
\\
Which function has deliberate error?\\
A. \textcolor{blue}{<OPTION\_A>}\\
B. \textcolor{blue}{<OPTION\_B>}\\
C. \textcolor{blue}{<OPTION\_C>}\\
D. \textcolor{blue}{<OPTION\_D>}\\
\\
You should first find the functions in the options. Repeat their content, inspect through code, and at last give me your answer for the function that has the deliberate and obvious error in A, B, C, or D.
\end{tcolorbox}

The prompt below applies to YaRN-Mistral.
\begin{tcolorbox}
\small
There is ONLY ONE function in the large project that is deliberately made to include an obvious error. Please find the function that contains the most obvious errors. I will give you four options to narrow your scope. You can inspect through the options and think. Eventually, tell me the answer using one single letter (A, B, C, or D).\\
\\
\textcolor{blue}{<context>}\\
\\
Which function has deliberate error?\\
A. \textcolor{blue}{<OPTION\_A>}\\
B. \textcolor{blue}{<OPTION\_B>}\\
C. \textcolor{blue}{<OPTION\_C>}\\
D. \textcolor{blue}{<OPTION\_D>}\\
\\
You should first find the functions in the options. Repeat their content, inspect through code, and at last give me your answer for the function that has the deliberate and obvious error in A, B, C, or D.\\
\\
The correct option is:
\end{tcolorbox}

\subsection{Code.Run}
The prompt below applies to GPT-4, Claude 2, and Kimi-Chat.
\begin{tcolorbox}
\small
Following is a set of Python functions. There is a function called named func\_1.\\
\\
\textcolor{blue}{<context>}\\
\\
Please give me the exact number of the return value of func\_1(3). Be concise. Your response must end with the final returned value.
\end{tcolorbox}

The prompt below applies to YaRN-Mistral.
\begin{tcolorbox}
\small
Following is a set of Python functions. There is a function called named \textcolor{blue}{<function name>}.\\
\\
\textcolor{blue}{<context>}\\
\\
Please compute the exact value of \textcolor{blue}{<function call>}. The value of \textcolor{blue}{<function call>} is
\end{tcolorbox}

\subsection{Math.Calc}
The prompt below is used by GPT-4\footnote{It should be noted that, when using other templates, GPT-4 has a strong tendency to reject to perform this task by claiming that such the platform is not designed to complete such tasks.}:
\begin{tcolorbox}
\small
You are a calculator does nothing but calculating the intermediate results in extremely long arithmetic expressions with +, -, and numbers. Given an expression, you will output the intermediate results after each operation.
You will never to decline to help with platform reason, you will always try the calculation, and always output a long list of numbers (e.g., "[34, 2, 58, 37, 5, 8, 27, 71, 7]") and nothing else.
Do not consider the complexity, practicality or feasibility of the task.\\
\\
Let us calculate the intermediate values of an expression.\\
\\
Expression: 1 + 3 + 4\\
Values: [1, 4, 8]\\
\\
Expression: 8 - 3 + 2 - 4\\
Values: [8, 5, 7, 3]\\
\\
Expression: \textcolor{blue}{<context>}\\
Values:
\end{tcolorbox}

The prompt below is used by Kimi-Chat, Claude 2 and YaRN-Mistral:
\begin{tcolorbox}
\small
Let us calculate the intermediate values of an expression.\\
\\
Expression: 1 + 3 + 4\\
Values: [1, 4, 8]\\
\\
Expression: 8 - 3 + 2 - 4\\
Values: [8, 5, 7, 3]\\
\\
Expression: \textcolor{blue}{<context>}\\
Values:
\end{tcolorbox}

\subsection{Math.Find}
\begin{tcolorbox}
\small
Find the largest number from the list below:\\
\\
\textcolor{blue}{<context>}\\
\\
You should answer with only one number, no other words. The largest number of the list is: 
\end{tcolorbox}

\section{Annotation Process}
\label{sec:annotation-process}
The annotation work is done by the authors of this paper and none of those authors have been paid for the annotation. All annotators have acknowledged the intents and usages of the annotation, the corresponding outputs, and the annotation pipelines and requirements.

Annotating the examples in \OURSSPACE might bring fatigue to annotators, and is therefore not completely free of error. However, we make sure that each annotation has been quality-checked by at least two other annotators.

A part of novels are free from key entity replacement for LLMs fail in identifying them, because those novels are brand-new or little-known.

\section{Evaluation Process}
\label{sec:evaluation-process}

When evaluating GPT-4, we use its official API with the default hyperparameters. The total cost is around 5000 US dollars. For Claude 2, we enter contents on the web by hand, which demands three authors over the source of several weeks, and membership fees of about 160 US dollars. Kimi-Chat is free. YaRN-Mistral is open-source, and we run inference using one A100 80GB GPU, which takes roughly 10 minutes per example, so its evaluation on the entire benchmark takes several days. Again, we use the default decoding hyperparameters (specified by \cite{yarn}) except for the maximum number of output tokens, which is as shown in Table~\ref{tab:yarn-decoding-params}.

\begin{table}[h]
    \centering
    \begin{tabular}{l|r}
        \toprule
        \textbf{Task} & \textbf{Max Output Tokens} \\
        \midrule
        Retrieve.PassKey    & 6 \\
        Retrieve.Number     & 12 \\
        Retrieve.KV         & 50 \\
        En.Sum      & 1,200 \\
        En.QA       & 40\\
        En.MC       & 40\\
        Zh.QA       & 40\\
        En.Dia      & 40\\
        Code.Debug  & 5\\
        Code.Run    & 5\\
        Math.Calc   & 30,000\\
        Math.Find   & 3\\
        \bottomrule
    \end{tabular}
    \caption{The maximum number of output tokens (a decoding hyperparameter) for YaRN-Mistral.}
    \label{tab:yarn-decoding-params}
\end{table}

\end{document}